\documentclass[11pt]{article}
\usepackage[final]{acl}

\usepackage{times}
\usepackage{latexsym}

\usepackage[T1]{fontenc}

\usepackage[utf8]{inputenc}

\usepackage{fontawesome5}

\usepackage{microtype}

\usepackage{inconsolata}

\usepackage{graphicx}

\usepackage{amsmath}
\usepackage{xspace}
\usepackage{amsfonts}
\usepackage{amssymb}
\usepackage{algorithm}
\usepackage{algpseudocode}
\usepackage[most]{tcolorbox}
\usepackage{tabularx}

\usepackage{booktabs}
\usepackage{multirow}
\usepackage{xcolor}
\usepackage{colortbl}
\usepackage{array}
\usepackage{makecell}
\usepackage{siunitx}
\usepackage{subcaption}
\usepackage{pifont}
\usepackage{enumitem}
\usepackage{bbm}
\usepackage{listings}

\usepackage{amsthm}
\usepackage{bm}

\newcounter{prompt}[section]
\renewcommand{\theprompt}{\thesection.\arabic{prompt}}

\newenvironment{promptbox*}[2][]{%
    \begin{figure*}[t]
    \refstepcounter{prompt}%
    \begin{tcolorbox}[
        colback=blue!5!white,
        colframe=blue!75!black,
        fonttitle=\bfseries,
        title={Prompt \theprompt: #2},
        enhanced,
        width=\textwidth,
        attach boxed title to top left={
            yshift=-2mm,
            xshift=5mm
        },
        boxed title style={
            colback=blue!75!black
        },
        sharp corners=south,
        drop shadow,
        left=1mm,
        right=1mm,
        top=2mm,
        bottom=1mm,
        boxsep=1mm,
        #1
    ]%
    \ttfamily\small
    \linespread{1.2}\selectfont
    \obeylines
    \obeyspaces
    \ignorespaces
}{%
    \end{tcolorbox}
    \end{figure*}
}

\newcommand{\sysname}{\textsc{DeepRewind}}

\title{\sysname{}: Predicting and Repairing Premature Commitments in Deep Research Agents}

\author{
Amirhossein Abaskohi\thanks{Corresponding author: \texttt{aabaskoh@cs.ubc.ca}.},
Amirhossein Dabiriaghdam, \\
\textbf{Lele Wang,
Peter West,
Giuseppe Carenini}\\[0.5em]
\\
University of British Columbia
}

\begin{document}
\maketitle

\begin{abstract}
Deep-research agents conduct long-horizon investigations through iterative search, evidence evaluation, belief revision, and synthesis. However, they may commit to claims before sufficient evidence is available, causing later reasoning to reinforce an incorrect interpretation. We introduce \textbf{\sysname{}}, an additive control layer for reversible deep research that represents the agent's evolving epistemic state as a typed graph of sources, evidence, claims, hypotheses, assumptions, commitments, plans, and drafts. Before accepting an intermediate conclusion, a prompt-based world model predicts its impact and estimates reversibility based on hypothesis narrowing, information loss, recovery cost, and contradiction-trigger coverage. A binary controller blocks risky commitments, while a consistency monitor performs dependency-aware rollback when later evidence invalidates them. Across DRBench and LiveDRBench, \sysname{} improves insight recall by \textbf{3.6 percentage points} and reduces premature commitments by \textbf{59.1\%} relative to Open Deep Research\footnote{Code is available on \href{https://github.com/AmirAbaskohi/DeepRewind}{{\textcolor{black}{\faGithub}}~GitHub}.}.
\end{abstract}


\section{Introduction}
\label{sec:introduction}

Deep research agents extend large language models (LLMs) from single-turn question answering to long-horizon information seeking, where an agent repeatedly plans searches, navigates the web, evaluates sources, reconciles conflicting evidence, and produces a citation-supported report~\citep{huang2025deepresearchagentssystematic,java2026characterizing}. Recent systems such as DeepResearcher and WebThinker demonstrate that integrating reasoning with real-world web interaction can substantially improve open-ended research capabilities~\citep{zheng-etal-2025-deepresearcher,li2025webthinker}. Correspondingly, benchmarks have begun to evaluate these agents on realistic, multi-source research tasks that require both comprehensive evidence gathering and coherent synthesis~\citep{du2026deepresearch,abaskohi2026drbench}. However, existing work primarily evaluates the quality of the final report, while paying less attention to how intermediate research decisions shape the trajectory.

A central failure mode is \emph{premature commitment}: the agent accepts a claim, favors a hypothesis, rejects an alternative, or drafts a conclusion before gathering sufficient evidence. Such commitments shape later queries, evidence selection, reasoning, and writing. Prior work shows that LLM agents may converge too early or become confident before completing the reasoning needed to justify their answers~\citep{mehta2026agentscommitsoondiagnosing,gai2026understandingmitigatingprematureconfidence}. Reflection, critique, tree search, and memory can revisit failures, explore alternatives, or preserve information~\citep{shinn2023reflexion,gou2024critic,zhou2024language,yao2023tree,xu2025amem,packer2024memgptllmsoperatingsystems}. However, they do not predict how a commitment changes the agent's epistemic state or identify the minimal downstream decisions to repair when later evidence invalidates it.

\begin{figure*}
    \centering
    \includegraphics[width=\linewidth]{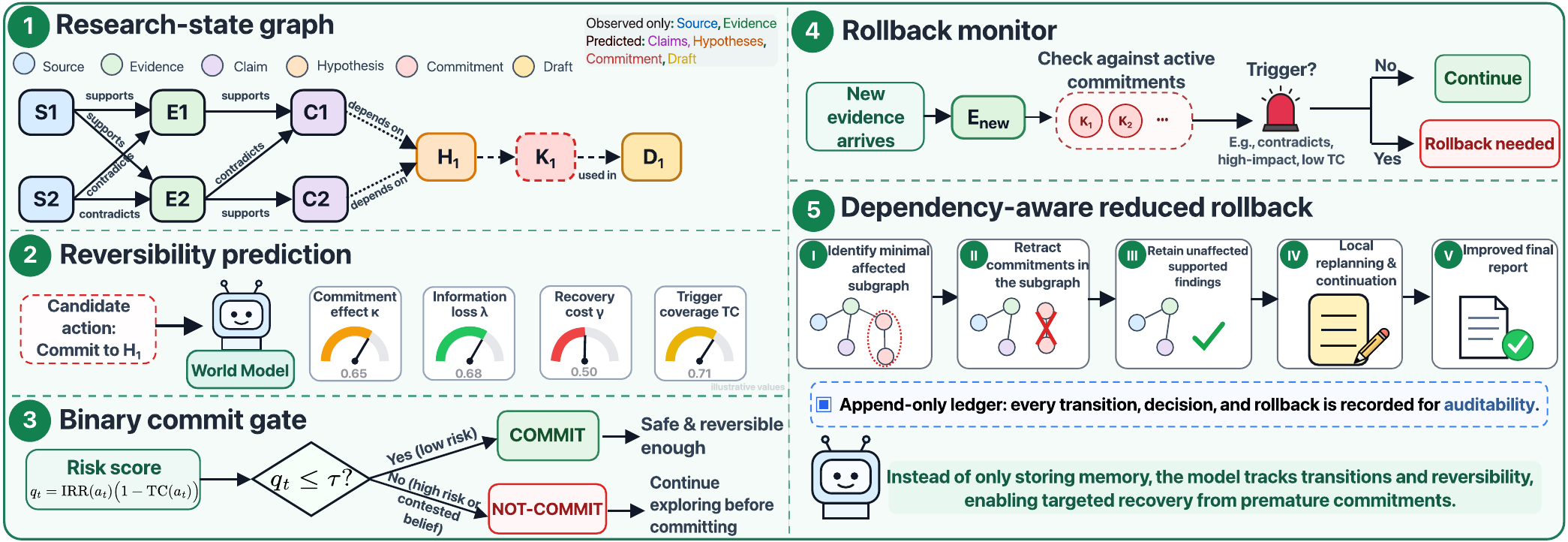}
    \caption{\textbf{Overview of \sysname{}.} \sysname{} models the agent's epistemic state as a typed graph, predicts commitment reversibility, and controls whether to commit. If new evidence invalidates a commitment, dependency-aware rollback removes only the affected subgraph while preserving independent findings.}
    \label{fig:deeprewind-method}
    \vspace{-1em}
\end{figure*}

We introduce \sysname{}, an additive control layer for reversible deep research built on Open Deep Research~\citep{langchain2025opendeepresearch}. It represents the agent's epistemic state as a typed graph of sources, evidence, claims, hypotheses, assumptions, commitments, plans, and drafts. Before accepting a commitment, a research-state world model predicts its graph effects and estimates reversibility from hypothesis narrowing, information loss, recovery cost, and contradiction likelihood. Inspired by graph world models and LLM-based web agents~\citep{feng2025graph,gu2025is,chae2025web}, \sysname{} models the agent's internal research state rather than the external environment. A binary controller blocks risky commitments, while a monitor performs dependency-aware rollback if evidence invalidates them.

We evaluate \sysname{} through controlled initial-condition and hypothesis-switching experiments, exposing agents to early supportive, contradictory, or distracting evidence before introducing evidence for a competing hypothesis. These experiments evaluate the downstream effectiveness of the overall control mechanism, including whether gating prevents lock-in and whether reduced rollback repairs commitments invalidated by later evidence. Across DRBench~\cite{abaskohi2026drbench} and LiveDRBench~\cite{java2026characterizing}, \sysname{} improves insight recall by \textbf{3.6 percentage points} and reduces premature commitments by \textbf{59.1\%} relative to Open Deep Research.

\section{Related Work}
\label{sec:related_work}

Recent work shows that LLMs and agents may commit to an answer before completing sufficient reasoning, causing later steps to reinforce the initial decision~\citep{mehta2026agentscommitsoondiagnosing,gai2026understandingmitigatingprematureconfidence}. Reflection and critique methods revise outputs after errors are observed~\citep{shinn2023reflexion,gou2024critic}, while tree-based methods explore alternative reasoning paths before selecting a solution~\citep{yao2023tree,zhou2024language}. However, they do not explicitly track which claims, assumptions, plans, and draft fragments depend on a commitment or determine what must be repaired when it fails. World models predict action consequences in structured or Web environments~\citep{feng2025graph,gu2025is}, and reversibility-aware reinforcement learning estimates whether environmental actions can be undone~\citep{grinsztajn2021there}, but both focus on the external environment rather than the agent's epistemic state. Classical truth-maintenance and dependency-directed backtracking systems record belief justifications and retract inconsistent assumptions~\citep{DOYLE1979231,STALLMAN1977135}, but do not predict the risk of LLM-generated commitments before they are made. \sysname{} models the research state as a typed graph, predicts how commitments narrow hypotheses, suppress conflicting information, and increase recovery costs, and performs dependency-aware reduced rollback when evidence invalidates them.

\section{\sysname{}}
\label{sec:method}

Figure~\ref{fig:deeprewind-method} provides an overview of \sysname{}, an additive control layer built on Open Deep Research~\citep{langchain2025opendeepresearch}. Without altering the base agent's search, tool use, or report generation, \sysname{} intervenes before intermediate conclusions are accepted, predicts their effect on the research state, estimates reversibility, and permits or blocks the commitment. Inspired by graph world models~\citep{feng2025graph} and WebDreamer~\citep{gu2025is}, it models transitions in the agent's internal epistemic state---including claims, hypotheses, assumptions, plans, commitments, and drafts---rather than changes to the external environment.

\paragraph{Research-state representation.}
At research step $t$, the agent's epistemic state is represented as a typed, weighted, directed graph $G_t=(V_t,E_t)$. Its nodes include \texttt{Source}, \texttt{Evidence}, \texttt{Claim}, \texttt{Hypothesis}, \texttt{Assumption}, \texttt{Commitment}, \texttt{DraftFragment}, and \texttt{PlanStep}; edges represent relations such as \texttt{supports}, \texttt{contradicts}, \texttt{depends\_on}, \texttt{used\_in}, and \texttt{locks\_in}. We distinguish externally observed \texttt{Source} and \texttt{Evidence} nodes from endogenous nodes produced by the agent. The world model may reference observed nodes but may only create or modify endogenous ones. State changes are recorded as graph deltas, $G_{t+1}=G_t\oplus\Delta G_t$, in an append-only ledger. Invalidated artifacts are marked contested, stale, or retracted rather than deleted, preserving the full research history. Following \citet{fatemi2024talk}, we serialize graph nodes, attributes, and typed relations into structured text.

\paragraph{Research-state world model.}
Before a candidate commitment $a_t$ is executed, the serialized research-state graph and candidate action are provided to the world model, which predicts:
\begin{equation}
\small
M_\phi(G_t,a_t)=
\left(
\widehat{\Delta G}_t,
\widehat{\bm m}_t
\right),
\label{eq:world_model}
\end{equation}
where $\widehat{\Delta G}_t$ contains predicted node and edge changes and $\widehat{\bm m}_t=(\hat{\kappa}_t,\hat{\lambda}_t,\hat{\gamma}_t,\hat{\theta}_t,\widehat{\mathrm{TC}}_t)$ contains reversibility metadata. Here, $\kappa$ measures how strongly the commitment narrows competing hypotheses, $\lambda$ estimates information loss, $\gamma$ estimates recovery cost, $\theta$ specifies the contradiction conditions that should invalidate the commitment, and $\mathrm{TC}$ estimates whether future research steps are likely to expose such contradictions. The prompt-based implementation emits these predictions as structured JSON. The target proposition, initial alternatives, and commitment transition are elicited using the prompts described in Appendix~\ref{app:prompts}.

\paragraph{Reversibility-aware commitment.}
From the current graph snapshot, \sysname{} computes claim beliefs using reliability-weighted supporting and contradicting evidence and converts claim support into a plausibility distribution over active hypotheses. Commitment effect is computed from reduced hypothesis entropy, information loss from the contradiction mass suppressed by the commitment, and recovery cost from the downstream endogenous nodes that would need to be retracted or regenerated. These quantities are combined as:
\begin{equation}
\begin{aligned}
\mathrm{IRR}(a_t)
&=
\operatorname{clip}_{[0,1]}
\left(
\alpha_\kappa\kappa_t+
\alpha_\lambda\lambda_t+
\alpha_\gamma\gamma_t
\right),\\
U(a_t)
&=
V(a_t)-
\eta\,\mathrm{IRR}(a_t)
\left(1-\mathrm{TC}(a_t)\right),
\end{aligned}
\label{eq:main_reversibility}
\end{equation}
where $V(a_t)$ is the estimated research value of the commitment. The controller outputs either \textsc{commit} or \textsc{not-commit}. Contested claims are never committed. In threshold mode, the action is committed only when $\mathrm{IRR}(a_t)(1-\mathrm{TC}(a_t))$ is below a configurable threshold; in utility mode, it is committed only when $U(a_t)$ exceeds a configurable threshold. A \textsc{not-commit} decision leaves the base agent free to gather more evidence, preserve competing hypotheses, or reformulate the claim.

\paragraph{Monitoring and reduced rollback.}
Each accepted commitment stores a trigger defined by minimum evidence reliability and contradiction strength. As new evidence enters the graph, a read-only monitor checks whether the trigger is met and the claim belief falls below a threshold. If so, \sysname{} deterministically retracts the commitment and its unique justifications, removes commitment-specific dependencies, and marks downstream hypotheses, plans, and drafts as stale or retracted. Shared or independently supported findings are preserved, limiting repair to the affected subgraph without additional LLM calls. This online monitor is part of the control mechanism and is distinct from the premature-commitment evaluation metric: rollback may occur as soon as the invalidation condition is satisfied, whereas premature commitment is measured retrospectively from the complete run trace using the same criterion for all methods.

\paragraph{Controlled interventions and logging.}
For controlled analysis, \sysname{} includes default-off hooks for initial-condition seeding and alternative switching. Seeding injects provenance-tagged supportive, contradictory, neutral, or distracting evidence, while switching first supports one alternative and later introduces evidence favoring a competitor. When disabled, these hooks do not affect the base agent. All predictions, decisions, triggers, and rollbacks are stored in append-only JSONL logs with configuration snapshots. Appendix~\ref{app:method_details} details the formulations, policies, interventions, rollback procedure, orchestration, and logging schema.

\section{Experiments}
\label{sec:experiments}

\paragraph{Experimental setup.}
We implement \sysname{} on Open Deep Research~\citep{langchain2025opendeepresearch}, keeping its search and report-generation workflow fixed. GPT-4o~\citep{openai2024gpt4technicalreport} serves as the research-state world model at temperature $0$, while base agents use temperature $0.7$ for up to 10 iterations. Model calls use OpenRouter\footnote{\url{https://openrouter.ai/}}, and Tavily\footnote{\url{https://www.tavily.com/}} retrieves up to five advanced-search results per query. We set $T=1$, $\alpha_\kappa=\alpha_\lambda=\alpha_\gamma=\tfrac{1}{3}$, $\eta=1$, $\tau_{\mathrm{commit}}=0.35$, $\beta^\star=0.5$, $\rho^\star=0.75$, $w^\star=0.6$, fallback $\mathrm{TC}=0.5$, and recovery budget $B=20$. We run each configuration with three random seeds and report the mean and standard deviation across runs; full configurations are provided in Appendix~\ref{app:experimental_details}. Across runs, \sysname{} invokes the world model \textbf{4.8} times per task on average, adding \textbf{6.5\%} input tokens, \textbf{4.2\%} output tokens, and \textbf{14\%} wall-clock latency relative to \textsc{ODR}.

\paragraph{Benchmarks and models.}
We evaluate on DRBench~\citep{abaskohi2026drbench} and LiveDRBench~\citep{java2026characterizing}, each with 100 tasks, reporting insight-level precision and recall against reference insights. We test GPT-4o~\citep{openai2024gpt4technicalreport}, GPT-5.4~\citep{openai2026gpt54}, Qwen3.5-27B~\citep{qwen3.5}, and DeepSeek-V4 Pro~\citep{deepseekai2026deepseekv4highlyefficientmilliontoken} as the LLM backbone of the agent. GPT-4o serves as the fixed world model, and all experiments use three random seeds. Appendix~\ref{app:world_model_ablation} evaluates \sysname{}'s generalizability across different LLM world models.

\subsection{Overall Performance}
\label{sec:overall_performance}

Table~\ref{tab:main_results} shows that \sysname{} consistently improves both precision and recall across all models and benchmarks. On average, it increases recall by \textbf{3.6 points} and precision by \textbf{3.1 points}, recovering more reference insights without adding unsupported content. GPT-5.4 achieves the strongest overall results, with \sysname{} reaching \textbf{40.3\%} recall on DRBench and \textbf{35.6\%} on LiveDRBench. A component ablation in Appendix~\ref{app:component_ablation} isolates the contributions of graph tracking, world-model prediction, pre-commitment gating, and reduced rollback. Table~\ref{tab:main_results_std} in Appendix~\ref{app:error} reports variation across runs, and Appendix~\ref{app:qualitative} provides a qualitative analysis.

\begin{table}[t]
\centering
\small
\caption{Insight precision and recall (\%) on DRBench and LiveDRBench, averaged across three runs.}
\label{tab:main_results}
\resizebox{\columnwidth}{!}{
\begin{tabular}{llcccc}
\toprule
\multirow{2}{*}{\textbf{Model}}
& \multirow{2}{*}{\textbf{Method}}
& \multicolumn{2}{c}{\textbf{DRBench}}
& \multicolumn{2}{c}{\textbf{LiveDRBench}} \\
\cmidrule(lr){3-4}
\cmidrule(lr){5-6}
& & \textbf{Prec.} & \textbf{Rec.}
& \textbf{Prec.} & \textbf{Rec.} \\
\midrule
GPT-4o
& \textsc{ODR}
& 55.8 & 16.2
& 51.4 & 13.8 \\
& \sysname{}
& \textbf{58.7} & \textbf{19.6}
& \textbf{54.6} & \textbf{17.1} \\
\midrule
GPT-5.4
& \textsc{ODR}
& 68.4 & 36.5
& 63.7 & 31.8 \\
& \sysname{}
& \textbf{71.2} & \textbf{40.3}
& \textbf{66.8} & \textbf{35.6} \\
\midrule
Qwen3.5-27B
& \textsc{ODR}
& 60.9 & 24.6
& 56.2 & 20.8 \\
& \sysname{}
& \textbf{63.8} & \textbf{28.1}
& \textbf{59.5} & \textbf{24.3} \\
\midrule
DeepSeek-V4 Pro
& \textsc{ODR}
& 64.7 & 29.8
& 59.8 & 25.4 \\
& \sysname{}
& \textbf{67.6} & \textbf{33.5}
& \textbf{63.1} & \textbf{29.0} \\
\bottomrule
\end{tabular}
}
\end{table}

\subsection{Robustness to Initial Evidence}
\label{sec:initial_condition_results}

We evaluate GPT-4o under neutral, supportive, contradictory, and distracting initial evidence. Resistance measures how often the agent avoids adopting injected claims before independent Web support. Table~\ref{tab:initial_condition_results} shows that all interventions degrade \textsc{ODR}, especially contradictory evidence. \sysname{} consistently preserves higher precision and recall, improving them by about \textbf{4.7} and \textbf{4.2 points} across non-neutral conditions. It also raises average resistance by \textbf{29.5 points}, with the largest gain of \textbf{37.8 points} under \textsc{Early-Counter}, showing greater robustness to misleading early evidence.

\begin{table}[t]
\centering
\small
\caption{Robustness of GPT-4o to initial evidence conditions (\%). Resistance measures the percentage of runs that avoid unsupported commitment to the injected position.}
\label{tab:initial_condition_results}
\resizebox{\columnwidth}{!}{
\begin{tabular}{llccccc}
\toprule
\textbf{Condition}
& \textbf{Method}
& \multicolumn{2}{c}{\textbf{DRBench}}
& \multicolumn{2}{c}{\textbf{LiveDRBench}}
& \textbf{Resist.} \\
\cmidrule(lr){3-4}
\cmidrule(lr){5-6}
& & \textbf{Prec.} & \textbf{Rec.}
& \textbf{Prec.} & \textbf{Rec.}
& \textbf{$\uparrow$} \\
\midrule
Neutral
& \textsc{ODR}
& 55.8 & 16.2
& 51.4 & 13.8
& -- \\
& \sysname{}
& \textbf{58.7} & \textbf{19.6}
& \textbf{54.6} & \textbf{17.1}
& -- \\
\midrule
Early-Support
& \textsc{ODR}
& 53.2 & 14.6
& 48.8 & 12.1
& 43.8 \\
& \sysname{}
& \textbf{58.0} & \textbf{18.8}
& \textbf{53.9} & \textbf{16.4}
& \textbf{72.4} \\
\midrule
Early-Counter
& \textsc{ODR}
& 52.1 & 13.5
& 47.6 & 11.0
& 38.2 \\
& \sysname{}
& \textbf{57.4} & \textbf{18.2}
& \textbf{53.1} & \textbf{15.8}
& \textbf{76.0} \\
\midrule
Early-Distractor
& \textsc{ODR}
& 54.7 & 15.3
& 50.2 & 12.9
& 61.2 \\
& \sysname{}
& \textbf{58.2} & \textbf{19.0}
& \textbf{54.1} & \textbf{16.6}
& \textbf{83.4} \\
\bottomrule
\end{tabular}
}
\end{table}

\subsection{Effect of Initial Belief Selection}
\label{sec:initial_belief_results}

We examine how initial beliefs shape GPT-4o's research trajectory and commitment behavior. \textsc{One-Belief} uses only the target proposition, while the alternative settings rank five competing beliefs by plausibility and select either the highest- or lowest-ranked one. This tests robustness to favorable and misleading starting hypotheses. Table~\ref{tab:initial_belief_results} shows that the strongest alternative improves both methods, whereas the weakest sharply degrades \textsc{ODR}, raising the fraction of accepted commitments that are later invalidated to \textbf{46.4\%}. \sysname{} remains stronger across all settings, with its largest gains under \textsc{Worst-Alternative}: about \textbf{6.5 points} in precision, \textbf{5.0 points} in recall, and a \textbf{26.2-point} reduction in premature commitments. The higher rollback rate reflects more frequent repair of unfavorable initial commitments.

\begin{table}[t]
\centering
\small
\caption{Effect of initial belief selection with GPT-4o (\%). PC is the percentage of accepted commitments that are later invalidated by subsequently collected evidence, and RB denotes rollback rate; rollback is not available for \textsc{ODR}.}
\label{tab:initial_belief_results}
\resizebox{\columnwidth}{!}{
\begin{tabular}{llcccccc}
\toprule
\multirow{2}{*}{\textbf{Initial belief}}
& \multirow{2}{*}{\textbf{Method}}
& \multicolumn{2}{c}{\textbf{DRBench}}
& \multicolumn{2}{c}{\textbf{LiveDRBench}}
& \textbf{PC}
& \textbf{RB} \\
\cmidrule(lr){3-4}
\cmidrule(lr){5-6}
& & \textbf{Prec.} & \textbf{Rec.}
& \textbf{Prec.} & \textbf{Rec.}
& \textbf{$\downarrow$}
& \textbf{$\downarrow$} \\
\midrule
One-Belief
& \textsc{ODR}
& 55.8 & 16.2
& 51.4 & 13.8
& 31.7 & -- \\
& \sysname{}
& \textbf{58.7} & \textbf{19.6}
& \textbf{54.6} & \textbf{17.1}
& \textbf{13.1} & 9.1 \\
\midrule
Best-Alternative
& \textsc{ODR}
& 57.1 & 17.5
& 52.6 & 15.0
& 24.0 & -- \\
& \sysname{}
& \textbf{59.1} & \textbf{20.5}
& \textbf{55.3} & \textbf{18.0}
& \textbf{8.5} & 6.3 \\
\midrule
Worst-Alternative
& \textsc{ODR}
& 50.3 & 12.9
& 46.1 & 10.4
& 46.4 & -- \\
& \sysname{}
& \textbf{56.8} & \textbf{17.9}
& \textbf{52.5} & \textbf{15.4}
& \textbf{20.2} & 15.1 \\
\bottomrule
\end{tabular}
}
\end{table}

\section{Conclusion}
\label{sec:conclusion}

We presented \sysname{}, an additive control layer for predicting, preventing, and repairing premature commitments in deep-research agents. It represents the evolving research state, blocks risky commitments before they influence later reasoning, and locally rolls back those invalidated by new evidence. Across DRBench and LiveDRBench, \sysname{} improves insight recall by \textbf{3.6 percentage points} and reduces premature commitments by \textbf{59.1\%} relative to Open Deep Research. Controlled experiments further show improved robustness to biased, distracting, and unfavorable initial states. These findings highlight the value of explicitly modeling commitment reversibility in long-horizon research. Future work may replace the prompt-based world model with learned graph or structured transition models.


\section*{Limitations}

\sysname{} depends on an LLM world model and an automatically constructed epistemic graph, so incorrect state extraction, missing dependencies, or poorly calibrated reversibility estimates may block useful commitments or fail to prevent harmful ones. We do not directly evaluate the accuracy or calibration of the world model's predicted graph transitions and reversibility quantities against ground truth, because DRBench and LiveDRBench do not provide annotations of the counterfactual research-state transition following a candidate commitment or gold values for hypothesis narrowing, information loss, recovery cost, and trigger coverage. Our experiments therefore evaluate the downstream utility of these predictions rather than their independent predictive accuracy. We evaluate only a small subset of hyperparameter configurations and select the best-performing setting, rather than conducting a comprehensive sensitivity analysis; we also do not evaluate how world-model prediction quality changes over the course of longer research trajectories. Constructing expert-annotated research-state transition data and studying these factors are important directions for future work. The method also depends on fixed thresholds and hand-designed relation types, and our experiments are limited to two deep-research benchmarks, four base models, and controlled evidence interventions.

\section*{Ethical Considerations}
\label{sec:ethics}

\sysname{} inherits the biases, hallucinations, and source-quality limitations of its underlying language models and Web retrieval system; an inaccurate world-model prediction may also create false confidence that a commitment is safe or recoverable. The method should therefore be used with source verification and human oversight in high-stakes domains. Our experiments use public benchmarks, models, and implementations under their respective licenses and involve no human subjects or newly collected personal data. We used LLM assistance only to improve manuscript presentation, with all technical decisions, experiments, and conclusions verified by the authors.


\bibliography{custom}


\appendix


\section{Implementation Details}
\label{app:method_details}

This appendix expands the research-state representation, graph serialization, structural scoring, commitment policy, monitoring procedure, rollback mechanism, controlled interventions, and artifact protocol summarized in Section~\ref{sec:method}. The prompts used to initialize and update the research state are provided in Appendix~\ref{app:prompts}.

\paragraph{Research-state graph.}
The research state at step $t$ is a typed, weighted, directed multigraph $G_t=(V_t,E_t)$. Each node $v\in V_t$ has a type $\tau(v)$ drawn from \texttt{Source}, \texttt{Evidence}, \texttt{Claim}, \texttt{Hypothesis}, \texttt{Assumption}, \texttt{Commitment}, \texttt{DraftFragment}, and \texttt{PlanStep}. Each edge $e=(u,v)\in E_t$ has a typed relation, including \texttt{supports}, \texttt{contradicts}, \texttt{derived\_from}, \texttt{compresses}, \texttt{depends\_on}, \texttt{used\_in}, and \texttt{locks\_in}, together with a weight $w_e\in[0,1]$. A research action produces a local delta $\Delta G_t=(\Delta V_t,\Delta E_t)$:
\begin{equation}
G_{t+1}=G_t\oplus\Delta G_t.
\label{eq:app_graph_update}
\end{equation}
\texttt{Source} and \texttt{Evidence} nodes are observational and may not be generated or rewritten by the world model. All other node types are endogenous and may be created, updated, contested, marked stale, or retracted. Updates are append-only and preserve previous node versions and provenance.

To communicate $G_t$ to the prompt-based world model, we follow the graph-to-text prompting strategy of \citet{fatemi2024talk}. Each node is serialized as a structured record containing its stable alias, type, status, and relevant attributes, while each edge is represented as a typed source--relation--target record with its weight. Stable aliases allow the model to reference existing nodes without reproducing their full content and make predicted updates directly mappable to the graph ledger. The candidate commitment is appended as a separate structured record:
\begin{equation}
x_t=
\operatorname{Serialize}(G_t)
\,\Vert\,
\operatorname{Serialize}(a_t),
\label{eq:app_graph_serialization}
\end{equation}
where $\Vert$ denotes textual concatenation.

\paragraph{Claim belief and hypothesis plausibility.}
For claim $c$, let $\mathrm{Supp}(c)$ and $\mathrm{Con}(c)$ denote its supporting and contradicting evidence edges. We define their reliability-weighted masses as
\begin{align}
M_{\mathrm{supp}}(c)
&=
\sum_{e\in\mathrm{Supp}(c)}
w_e\rho_e,
\label{eq:app_support_mass}\\
M_{\mathrm{con}}(c)
&=
\sum_{e\in\mathrm{Con}(c)}
w_e\rho_e,
\label{eq:app_contra_mass}
\end{align}
where $\rho_e\in[0,1]$ is inherited from the source associated with evidence edge $e$. Claim belief is
\begin{equation}
\beta_t(c)=
\sigma\left(
\beta_0+
M_{\mathrm{supp}}(c)-
M_{\mathrm{con}}(c)
\right).
\label{eq:app_claim_belief}
\end{equation}
A hypothesis $h$ aggregates belief from its supporting claims:
\begin{align}
g_t(h)
&=
\ell_t(h)+
\sum_{c\rightarrow h}
w_{c,h}\beta_t(c),
\label{eq:app_hypothesis_score}\\
p_t(h)
&=
\frac{\exp(g_t(h)/T)}
{\sum_{h'\in\mathcal{H}_t}
\exp(g_t(h')/T)},
\label{eq:app_hypothesis_prob}
\end{align}
where $\ell_t(h)$ represents existing lock-in, $\mathcal{H}_t$ is the set of active hypotheses, and $T>0$ is a temperature parameter.

\paragraph{World-model interface.}
The implementation supplies the serialized input $x_t$ to the prompt-based world model. For clarity, we abstract this operation as conditioning directly on the graph and candidate action:
\begin{equation}
\begin{aligned}
M_\phi(G_t,a_t)
&=
\left(
\widehat{\Delta G}_t,
\widehat{\bm m}_t
\right),\\
\widehat{\bm m}_t
&=
\left(
\hat{\kappa}_t,
\hat{\lambda}_t,
\hat{\gamma}_t,
\hat{\theta}_t,
\widehat{\mathrm{TC}}_t
\right).
\end{aligned}
\label{eq:app_world_model}
\end{equation}
The structured prediction contains node creations or updates, edge additions or removals, reversibility estimates, an optional rollback trigger, and optional trigger coverage. Predicted aliases are resolved against the serialized graph before any update is applied. Existing \texttt{Source} and \texttt{Evidence} aliases may appear as edge endpoints, but the model may not create, modify, or reproduce their content. The complete input and output formats are given in Appendix~\ref{app:prompts}.

\paragraph{Commitment effect.}
Let $p_t$ be the current hypothesis distribution and $\hat{p}_{t+1}$ the distribution obtained after applying the predicted graph delta. The commitment effect measures normalized entropy reduction:
\begin{align}
H(p)
&=
-\sum_{h\in\mathcal{H}_t}
p(h)\log p(h),
\label{eq:app_entropy}\\
\kappa_t
&=
\operatorname{clip}_{[0,1]}
\left(
\frac{
H(p_t)-H(\hat{p}_{t+1})
}{
\log K
}
\right),
\label{eq:app_kappa}
\end{align}
where $K=|\mathcal{H}_t|$. A large value indicates that the candidate commitment sharply concentrates belief on a small subset of alternatives.

\paragraph{Information loss.}
For the claim targeted by the commitment, information loss is approximated by the fraction of reliability-weighted evidence mass that contradicts the committed position:
\begin{equation}
\lambda_t=
\operatorname{clip}_{[0,1]}
\left(
\frac{
M_{\mathrm{con}}(c)
}{
M_{\mathrm{supp}}(c)+
M_{\mathrm{con}}(c)
}
\right).
\label{eq:app_lambda}
\end{equation}
When the denominator is zero, we set $\lambda_t=0$.

\paragraph{Recovery cost.}
Let $D(\widehat{\Delta V}_t)$ denote the predictable endogenous nodes that transitively depend on nodes introduced or modified by the commitment. Recovery cost is
\begin{equation}
\gamma_t=
\operatorname{clip}_{[0,1]}
\left(
\frac{
|D(\widehat{\Delta V}_t)|
}{
B
}
\left(
c_{\mathrm{ret}}+
c_{\mathrm{regen}}
\right)
\right),
\label{eq:app_gamma}
\end{equation}
where $c_{\mathrm{ret}}$ and $c_{\mathrm{regen}}$ are the unit costs of retracting and regenerating a dependent node and $B$ is the recovery budget.

\paragraph{Rollback trigger and trigger coverage.}
For a commitment targeting claim $c$, the predicted trigger is
\begin{equation}
\theta_c(e)=
\mathbf{1}\left[
\begin{aligned}
&e\in\mathrm{Con}(c),\\
&\rho_e\geq\rho^\star,\\
&w_e\geq w^\star
\end{aligned}
\right].
\label{eq:app_trigger}
\end{equation}
Trigger coverage $\mathrm{TC}(a_t)\in[0,1]$ estimates the likelihood that future research probes expose an invalidating contradiction. It may be supplied by a constant fallback, predicted by the LLM, or estimated structurally from open or pending plan steps linked to the committed claim.

\paragraph{Irreversibility risk and utility.}
The three reversibility quantities are combined as
\begin{equation}
\mathrm{IRR}(a_t)=
\operatorname{clip}_{[0,1]}
\left(
\alpha_\kappa\kappa_t+
\alpha_\lambda\lambda_t+
\alpha_\gamma\gamma_t
\right),
\label{eq:app_irr}
\end{equation}
where $\alpha_\kappa,\alpha_\lambda,\alpha_\gamma\geq0$ and
\begin{equation}
\alpha_\kappa+
\alpha_\lambda+
\alpha_\gamma=1.
\label{eq:app_irr_weights}
\end{equation}
The utility-aware score is
\begin{equation}
U(a_t)=
V(a_t)-
\eta\,\mathrm{IRR}(a_t)
\left(
1-\mathrm{TC}(a_t)
\right),
\label{eq:app_utility}
\end{equation}
where $V(a_t)$ is configurable, such as zero or the predicted plausibility gain of the target hypothesis.

\paragraph{Commitment policy.}
Let
\begin{equation}
q_t=
\mathrm{IRR}(a_t)
\left(
1-\mathrm{TC}(a_t)
\right).
\label{eq:app_monitored_risk}
\end{equation}
The threshold-based policy is
\begin{equation}
\pi_{\mathrm{thr}}(a_t)=
\begin{cases}
\textsc{not-commit},
& \mathrm{Contested}(a_t),\\
\textsc{commit},
& q_t\leq\tau_{\mathrm{commit}},\\
\textsc{not-commit},
& \text{otherwise}.
\end{cases}
\label{eq:app_threshold_policy}
\end{equation}
The utility-based alternative is
\begin{equation}
\pi_{\mathrm{util}}(a_t)=
\begin{cases}
\textsc{not-commit},
& \mathrm{Contested}(a_t),\\
\textsc{commit},
& U(a_t)\geq\tau_U,\\
\textsc{not-commit},
& \text{otherwise}.
\end{cases}
\label{eq:app_utility_policy}
\end{equation}
A claim is considered contested when it has substantial support and contradiction mass and its current belief lies near the decision boundary.

\paragraph{Consistency monitoring.}
After a commitment becomes active, the monitor checks it whenever new evidence is added. A commitment on claim $c$ fires when qualifying contradiction evidence exists and the current claim belief falls below $\beta^\star$:
\begin{equation}
\begin{aligned}
\mathrm{Fire}(c)=1
\Longleftrightarrow {}&
\exists e\in\mathrm{Con}(c)
\text{ such that}\\
&\rho_e\geq\rho^\star,\quad
w_e\geq w^\star,\\
&\beta_{\mathrm{now}}(c)<\beta^\star.
\end{aligned}
\label{eq:app_fire}
\end{equation}
The monitor does not modify the graph. It returns the fired commitment, current claim belief, qualifying evidence identifiers, and the thresholds that were crossed.

\paragraph{Deterministic reduced rollback.}
For each fired commitment, \sysname{} first retracts the commitment node and any justification unique to it. Independently supported or shared justifications are retained. It then removes commitment-specific dependency edges and processes downstream nodes according to their type: dependent \texttt{DraftFragment}, \texttt{PlanStep}, \texttt{Hypothesis}, and \texttt{Assumption} nodes are marked stale, contested, or retracted according to the configured repair policy. The target claim is reopened when additional investigation is possible. Every operation is appended to the ledger as a rollback event, and no additional language-model call is required.

\paragraph{Initial-condition interventions.}
The initial-condition hook can insert provenance-tagged synthetic evidence before the supervisor begins reasoning. The supported conditions are \texttt{neutral}, which injects nothing; \texttt{support}, which supports the target proposition; \texttt{counter}, which contradicts it; and \texttt{distract}, which introduces low-weight peripheral evidence. Synthetic nodes record their condition, generation reason, and synthetic provenance so that they remain distinguishable from Web-derived evidence.

\paragraph{Alternative-switching interventions.}
The switching hook begins with two mutually distinguishable alternatives $A$ and $B$. It may insert early evidence supporting $A$, followed later by high-confidence contradiction to the committed-$A$ claim and high-confidence support for $B$. The switch can occur after the first accepted commitment or after a configurable number of research steps. When the hook is disabled, no synthetic evidence or switching behavior is introduced.

\paragraph{Experiment orchestration.}
The experiment driver enumerates runs over
\begin{equation}
\mathcal{R}=
\mathcal{Q}\times
\mathcal{F}\times
\mathcal{C}\times
\mathcal{A}\times
\mathcal{S},
\label{eq:app_experiment_grid}
\end{equation}
where $\mathcal{Q}$ contains research questions, $\mathcal{F}$ intervention families, $\mathcal{C}$ conditions, $\mathcal{A}$ experimental arms, and $\mathcal{S}$ random seeds. Each cell receives an isolated run directory and an arm-specific configuration. Completed cells are identified through marker files and skipped during resumed execution. Failures are logged with their traceback without stopping the remaining runs. Benchmark-specific dataset loading is kept outside the control layer.

\paragraph{Logging and reproducibility.}
World-model artifacts are stored as append-only JSONL records. Event types include \texttt{run\_start}, prediction events, policy decisions, \texttt{monitor\_check}, \texttt{trigger\_fired}, rollback events, and \texttt{run\_end}. The initial record stores the complete runtime configuration, including thresholds, coefficients, trigger-coverage mode, rollback policy, and intervention flags. Reproducibility is further supported through deterministic rollback planning, isolated run directories, manifest tracking, completion markers, and default-off intervention hooks.

\section{Experimental Details}
\label{app:experimental_details}

This appendix provides the model, agent, retrieval, control, evaluation, and reproducibility configurations used in Section~\ref{sec:experiments}.

\paragraph{Model configuration.}
All language-model requests are routed through OpenRouter using the canonical identifiers \texttt{openai/gpt-4o}, \texttt{openai/gpt-5.4}, \texttt{qwen/qwen3.5-27b}, and \texttt{deepseek/deepseek-v4-pro}. The base research models use temperature $0.7$, top-$p$ $1.0$, and a shared input-context limit of $128{,}000$ tokens. Intermediate research generations are limited to $4{,}096$ output tokens, while final report generation is limited to $16{,}384$ tokens. For models exposing an explicit reasoning-effort parameter, we use the medium setting; otherwise, we retain the provider default. We use the default balanced OpenRouter routing policy, allow provider-level retries, and do not permit fallback to a different model. Each failed model call is retried at most three times with exponential backoff of $2$, $4$, and $8$ seconds.

GPT-4o is used as the research-state world model in every experiment. We set its temperature to $0$, top-$p$ to $1$, and maximum output length to $4{,}096$ tokens. Structured JSON output is requested whenever supported by the provider. Invalid JSON responses are repaired through one deterministic parsing pass and, if parsing still fails, the prediction call is repeated up to two times. A candidate commitment is rejected when no valid prediction is obtained after these attempts.

\paragraph{Open Deep Research configuration.}
We retain the standard supervisor--researcher organization of Open Deep Research and apply the same configuration to all base models. Each run permits at most $10$ supervisor iterations, at most $3$ concurrent research units, and at most $5$ Web searches per research unit. Each research unit receives the complete research brief and the findings accumulated at the time it is invoked. Research terminates when the supervisor emits its completion decision or exhausts the iteration budget. The final report is generated from all accepted findings and active research-state artifacts. The baseline and \sysname{} conditions use identical prompts, iteration budgets, search tools, and report-generation limits; only the graph instrumentation and commitment-control operations differ.

\paragraph{Web retrieval.}
Web search is performed through Tavily with \texttt{search\_depth=advanced}, \texttt{topic=general}, \texttt{max\_results=5}, \texttt{include\_answer=false}, and \texttt{include\_raw\_content=true}. We retain up to three relevant text chunks from each returned page and cap the extracted content supplied to the agent at $8{,}000$ characters per source. No fixed domain allowlist or blocklist is applied. Duplicate URLs are removed within each run, but the same source may be retrieved by different experimental conditions. Search failures are retried twice before the corresponding query is recorded as unsuccessful.

\paragraph{\sysname{} configuration.}
The primary experiments use the threshold-based commitment policy. We set the hypothesis-softmax temperature to $T=1$, the reversibility weights to $\alpha_\kappa=\alpha_\lambda=\alpha_\gamma=\tfrac{1}{3}$, the risk penalty to $\eta=1$, and the commitment threshold to $\tau_{\mathrm{commit}}=0.35$. Trigger coverage is taken from the world-model prediction when available and otherwise defaults to $\mathrm{TC}=0.5$. We set the claim-belief trigger to $\beta^\star=0.5$, minimum source reliability to $\rho^\star=0.75$, and minimum contradiction-edge weight to $w^\star=0.6$. A claim is treated as contested when both its support and contradiction masses are at least $0.3$ and $\beta_t(c)\in[0.4,0.6]$.

The recovery budget is $B=20$, with unit retraction and regeneration costs $c_{\mathrm{ret}}=1$ and $c_{\mathrm{regen}}=2$. During rollback, unique commitment justifications are retracted, independently supported nodes are preserved, and downstream draft fragments and plan steps are marked stale rather than deleted. The utility-policy ablation uses
\begin{equation}
V(a_t)=
\max\left(
0,\,
\hat{p}_{t+1}(h^\star)-p_t(h^\star)
\right),
\label{eq:app_action_value}
\end{equation}
where $h^\star$ is the target hypothesis, and commits when $U(a_t)\geq\tau_U$ with $\tau_U=0$.

\paragraph{Compared conditions.}
We evaluate six conditions under identical research budgets. \textsc{ODR} is the unmodified Open Deep Research baseline. \textsc{Graph-Only} records the epistemic graph and transition ledger but does not invoke the world model or alter commitments. \textsc{Predict-Only} invokes the world model and records its risk estimates, but all candidate commitments are accepted and no rollback is performed. \textsc{No-Gate} accepts every candidate commitment but enables consistency monitoring and reduced rollback, isolating post-commitment repair. \textsc{\sysname{}} enables prediction, pre-commitment gating, monitoring, and rollback. We additionally evaluate \textsc{\sysname{}-NoRollback} to isolate the effect of pre-commitment gating without post-commitment repair. Results of this component ablation are reported in Appendix~\ref{app:component_ablation}.

For initial-condition experiments, supportive and contradictory synthetic evidence use reliability $0.85$ and edge weight $0.8$, while distracting evidence uses reliability $0.5$ and edge weight $0.25$. In switching experiments, early support for alternative $A$ uses reliability $0.8$ and weight $0.75$; the later contradiction to $A$ and support for $B$ use reliability $0.95$ and weight $0.9$. The primary switch occurs immediately after the first accepted commitment to $A$. All synthetic artifacts are explicitly tagged and excluded from source-quality analyses.

\paragraph{Evaluation.}
We use the released reference insights and official evaluation procedure for each benchmark. Let $\widehat{\mathcal{I}}_q$ be the insights extracted from the generated report for task $q$, $\mathcal{I}_q$ the reference insights, and $m_q$ the number of one-to-one matched insight pairs. We compute
\begin{equation}
\begin{aligned}
P_q&=\frac{m_q}{|\widehat{\mathcal{I}}_q|},\\
R_q&=\frac{m_q}{|\mathcal{I}_q|}.
\end{aligned}
\label{eq:app_insight_metrics}
\end{equation}
Precision is set to zero when no insight is produced. We first macro-average precision and recall across the $100$ tasks in each benchmark and then report the mean and standard deviation across three seeds. The evaluator receives only the generated report and reference insights and is not given the model identity or experimental condition.

We additionally report premature-commitment rate, commitment rate, blocked-commitment rate, trigger rate, rollback rate, average rollback size, and normalized recovery cost. Premature commitment is defined formally below and is evaluated from the complete run trace using the same criterion for all methods. We do not directly evaluate the accuracy of the world model's predicted graph transitions or reversibility estimates because the benchmarks do not provide ground-truth annotations for these quantities. Instead, we evaluate their utility through downstream performance, controlled interventions, component ablations, and sensitivity to the choice of world-model LLM. Synthetic intervention evidence is included when assessing resistance to premature commitment but excluded from final insight precision and recall unless it is independently supported by retrieved Web evidence.

\paragraph{Premature-commitment metric.}
We operationally define a premature commitment as an accepted commitment that is
subsequently invalidated by evidence collected later in the same research run.
For a commitment to claim $c$ made at step $t$, we examine all evidence retrieved
after $t$ and mark the commitment as premature if, by the end of the run, there
exists contradicting evidence satisfying
$\rho_e \geq \rho^\star$ and $w_e \geq w^\star$, and the resulting claim belief
falls below $\beta^\star$. Thus, the premature-commitment rate is

\begin{equation}
\mathrm{PC}
=
\frac{
\#\{\text{invalidated commitments}\}
}{
\#\{\text{accepted commitments}\}
}.
\label{eq:app_pc_metric}
\end{equation}

\begin{table*}[t]
\centering
\small
\caption{
Component ablation of \sysname{} with GPT-4o as the deep-research agent and
GPT-4o as the world model. Results are averaged across three runs.
PC denotes premature-commitment rate and RB denotes rollback rate.
}
\label{tab:component_ablation}
\resizebox{0.75\linewidth}{!}{
\begin{tabular}{lcccccc}
\toprule
\multirow{2}{*}{\textbf{Method}}
& \multicolumn{2}{c}{\textbf{DRBench}}
& \multicolumn{2}{c}{\textbf{LiveDRBench}}
& \textbf{PC}
& \textbf{RB} \\
\cmidrule(lr){2-3}
\cmidrule(lr){4-5}
& \textbf{Prec.} & \textbf{Rec.}
& \textbf{Prec.} & \textbf{Rec.}
& \textbf{$\downarrow$}
& \textbf{(\%)} \\
\midrule

\textsc{ODR}
& 55.8 & 16.2
& 51.4 & 13.8
& 31.7 & -- \\

\textsc{Graph-Only}
& 55.9 & 16.3
& 51.5 & 13.9
& 31.5 & -- \\

\textsc{Predict-Only}
& 56.1 & 16.5
& 51.7 & 14.1
& 30.9 & -- \\

\textsc{No-Gate}
& 57.2 & 17.8
& 52.9 & 15.4
& 31.2 & 14.7 \\

\textsc{\sysname{}-NoRollback}
& 58.2 & 19.0
& 54.0 & 16.5
& 14.5 & -- \\

\textsc{\sysname{}}
& \textbf{58.7} & \textbf{19.6}
& \textbf{54.6} & \textbf{17.1}
& \textbf{13.1} & 9.1 \\

\bottomrule
\end{tabular}
}
\end{table*}

The judgment is made retrospectively at the end of each run using the full set of
subsequently collected evidence, allowing the same criterion to be applied to
both \textsc{ODR} and \sysname{}. For \sysname{}, an online rollback may occur
earlier when the same invalidation condition is first satisfied, but the reported
PC metric is computed independently from the final run trace. If later evidence
remains mixed or ambiguous and does not satisfy both the contradiction and belief
thresholds, the commitment is not counted as premature.

\paragraph{Randomness and failure handling.}
We use seeds $13$, $42$, and $97$. Seeds control intervention construction, alternative ordering, concurrent-research scheduling, and the API seed parameter where supported. Because hosted APIs may remain nondeterministic, all aggregate results are computed from independently executed runs rather than replayed outputs. A run is considered failed only after three retries of its terminal error. Persistent failures receive zero precision and recall and remain included in the aggregate; we also report the completion rate for each model and condition.

\paragraph{Logging and resource accounting.}
Each run is written to an isolated directory containing its configuration, research-state ledger, world-model predictions, policy decisions, search results, monitor events, rollback records, generated report, evaluator output, and completion marker. The manifest records the benchmark, task identifier, model, condition, seed, execution status, and artifact paths. We record input, output, and reasoning-token counts when exposed by the provider, together with the number of model calls, Tavily searches, retrieved pages, research iterations, and wall-clock duration. At most four benchmark runs are executed concurrently, and completed cells are skipped when an interrupted experiment is resumed.

\section{Component Ablation}
\label{app:component_ablation}

We isolate the contribution of each component of \sysname{} using GPT-4o as both the deep-research agent and the research-state world model. All variants use the same research budget, retrieval configuration, prompts, and evaluation procedure. \textsc{Graph-Only} adds only the epistemic graph and transition logging, while \textsc{Predict-Only} additionally invokes the world model but does not use its predictions to alter commitments. \textsc{No-Gate} removes pre-commitment gating while retaining consistency monitoring and reduced rollback. \textsc{\sysname{}-NoRollback} retains prediction and gating but disables post-commitment rollback. Full \sysname{} combines prediction, gating, monitoring, and rollback.

Table~\ref{tab:component_ablation} isolates the roles of prediction, prevention, and repair. Graph construction and world-model prediction alone have limited effect when they do not alter the agent's decisions. Enabling reduced rollback improves final performance by repairing commitments after contradictory evidence is observed, while pre-commitment gating provides a larger gain by preventing risky commitments from propagating through the research trajectory. Combining gating with rollback yields the strongest overall precision and recall and the lowest premature-commitment rate.

\section{Error Analysis}
\label{app:error}

\begin{table*}[t]
\centering
\small
\caption{Mean $\pm$ standard deviation for the main results across three runs.}
\label{tab:main_results_std}
\resizebox{0.7\linewidth}{!}{
\begin{tabular}{llcccc}
\toprule
\multirow{2}{*}{\textbf{Model}}
& \multirow{2}{*}{\textbf{Method}}
& \multicolumn{2}{c}{\textbf{DRBench}}
& \multicolumn{2}{c}{\textbf{LiveDRBench}} \\
\cmidrule(lr){3-4}
\cmidrule(lr){5-6}
& & \textbf{Prec.} & \textbf{Rec.}
& \textbf{Prec.} & \textbf{Rec.} \\
\midrule
GPT-4o
& \textsc{ODR}
& $55.8{\pm}0.8$ & $16.2{\pm}0.5$
& $51.4{\pm}1.0$ & $13.8{\pm}0.7$ \\
& \sysname{}
& $\mathbf{58.7{\pm}0.6}$ & $\mathbf{19.6{\pm}0.4}$
& $\mathbf{54.6{\pm}0.8}$ & $\mathbf{17.1{\pm}0.6}$ \\
\midrule
GPT-5.4
& \textsc{ODR}
& $68.4{\pm}0.6$ & $36.5{\pm}0.7$
& $63.7{\pm}0.8$ & $31.8{\pm}0.8$ \\
& \sysname{}
& $\mathbf{71.2{\pm}0.5}$ & $\mathbf{40.3{\pm}0.6}$
& $\mathbf{66.8{\pm}0.6}$ & $\mathbf{35.6{\pm}0.7}$ \\
\midrule
Qwen3.5-27B
& \textsc{ODR}
& $60.9{\pm}0.9$ & $24.6{\pm}0.6$
& $56.2{\pm}1.0$ & $20.8{\pm}0.8$ \\
& \sysname{}
& $\mathbf{63.8{\pm}0.7}$ & $\mathbf{28.1{\pm}0.5}$
& $\mathbf{59.5{\pm}0.8}$ & $\mathbf{24.3{\pm}0.6}$ \\
\midrule
DeepSeek-V4 Pro
& \textsc{ODR}
& $64.7{\pm}0.7$ & $29.8{\pm}0.6$
& $59.8{\pm}0.9$ & $25.4{\pm}0.8$ \\
& \sysname{}
& $\mathbf{67.6{\pm}0.6}$ & $\mathbf{33.5{\pm}0.5}$
& $\mathbf{63.1{\pm}0.7}$ & $\mathbf{29.0{\pm}0.6}$ \\
\bottomrule
\end{tabular}
}
\end{table*}

\begin{table*}[t]
\centering
\small
\caption{
Paired statistical comparison of \sysname{} against \textsc{ODR}.
Differences are computed per task after averaging each task across three runs.
We report the mean improvement ($\Delta$), 95\% paired-bootstrap confidence
interval, and Holm-adjusted two-sided $p$-value over the 100 benchmark tasks.
}
\label{tab:significance}
\resizebox{0.75\linewidth}{!}{
\begin{tabular}{llcccc}
\toprule
\textbf{Model}
& \textbf{Benchmark}
& \textbf{Metric}
& \textbf{$\Delta$}
& \textbf{95\% CI}
& \textbf{$p_{\mathrm{adj}}$} \\
\midrule

GPT-4o
& DRBench & Precision
& +2.9 & [1.1, 4.7] & 0.012 \\
&         & Recall
& +3.4 & [1.8, 5.0] & 0.004 \\
& LiveDRBench & Precision
& +3.2 & [1.2, 5.1] & 0.011 \\
&             & Recall
& +3.3 & [1.5, 5.2] & 0.006 \\

\midrule
GPT-5.4
& DRBench & Precision
& +2.8 & [1.2, 4.4] & 0.009 \\
&         & Recall
& +3.8 & [2.1, 5.5] & 0.003 \\
& LiveDRBench & Precision
& +3.1 & [1.3, 4.9] & 0.008 \\
&             & Recall
& +3.8 & [1.9, 5.7] & 0.004 \\

\midrule
Qwen3.5-27B
& DRBench & Precision
& +2.9 & [0.9, 4.8] & 0.019 \\
&         & Recall
& +3.5 & [1.7, 5.3] & 0.006 \\
& LiveDRBench & Precision
& +3.3 & [1.0, 5.5] & 0.018 \\
&             & Recall
& +3.5 & [1.4, 5.6] & 0.009 \\

\midrule
DeepSeek-V4 Pro
& DRBench & Precision
& +2.9 & [1.1, 4.7] & 0.014 \\
&         & Recall
& +3.7 & [1.9, 5.5] & 0.005 \\
& LiveDRBench & Precision
& +3.3 & [1.1, 5.4] & 0.015 \\
&             & Recall
& +3.6 & [1.6, 5.6] & 0.007 \\

\bottomrule
\end{tabular}
}
\end{table*}

Table~\ref{tab:main_results_std} reports the run-level variation corresponding to Table~\ref{tab:main_results}. The relatively small standard deviations indicate that the gains from \sysname{} are consistent across the three independently executed runs rather than being driven by a single favorable trajectory. Variation is slightly higher on LiveDRBench, where changing Web content and retrieval results introduce additional uncertainty. Because three runs provide limited statistical power on their own, we additionally perform paired task-level comparisons between \sysname{} and \textsc{ODR}. Table~\ref{tab:significance} reports the mean per-task improvement, 95\% paired-bootstrap confidence intervals, and Holm-adjusted two-sided $p$-values across the 100 tasks in each benchmark.

Inspection of unsuccessful trajectories reveals three recurring error sources. First, we observe cases consistent with the world model underestimating the downstream effect of a commitment, particularly when its dependencies are only expressed implicitly in later draft fragments or plans. Second, highly reliable but incomplete evidence can produce a low predicted information-loss score, allowing the agent to commit before alternative explanations have been sufficiently explored. Third, rollback can preserve a stale dependent when that dependency is missing or incorrectly typed in the research-state graph. These failures suggest that the effectiveness of \sysname{} depends jointly on accurate reversibility prediction and sufficiently complete dependency recording.

\begin{table}[h]
\centering
\small
\caption{Ablation of the world-model LLM with GPT-4o fixed as the deep-research agent. Results are insight precision and recall (\%), averaged across three runs. GPT-4o is the default world model.}
\label{tab:world_model_ablation}
\resizebox{\columnwidth}{!}{
\begin{tabular}{lcccc}
\toprule
\multirow{2}{*}{\textbf{World model}}
& \multicolumn{2}{c}{\textbf{DRBench}}
& \multicolumn{2}{c}{\textbf{LiveDRBench}} \\
\cmidrule(lr){2-3}
\cmidrule(lr){4-5}
& \textbf{Prec.} & \textbf{Rec.}
& \textbf{Prec.} & \textbf{Rec.} \\
\midrule
Qwen3.5-27B
& 57.8 & 18.8
& 53.7 & 16.2 \\
DeepSeek-V4 Pro
& 58.4 & 19.2
& 54.2 & 16.8 \\
GPT-4o (default)
& 58.7 & 19.6
& 54.6 & 17.1 \\
GPT-5.4
& \textbf{59.2} & \textbf{20.1}
& \textbf{55.1} & \textbf{17.7} \\
\bottomrule
\end{tabular}
}
\end{table}

\section{World-Model LLM Ablation}
\label{app:world_model_ablation}

We examine whether \sysname{} depends on the LLM used to predict research-state transitions. The base deep-research agent is fixed to GPT-4o, and only the world-model LLM is changed. All models receive the same serialized research-state graph, commitment-prediction prompt, thresholds, and search budget. Table~\ref{tab:world_model_ablation} shows that stronger world models provide modest gains, but the overall performance is stable across model choices. GPT-5.4 achieves the highest precision and recall, while the default GPT-4o world model remains within one percentage point on most metrics. This experiment evaluates downstream sensitivity to the choice of world-model LLM rather than the accuracy or calibration of individual predicted transitions.

The limited variation indicates that the gains of \sysname{} do not arise solely from using a particular world-model LLM. Although GPT-5.4 produces the strongest results, its average recall improvement over GPT-4o is only $0.6$ percentage points. We therefore retain GPT-4o as the default world model in the main experiments.

\section{Qualitative Analysis}
\label{app:qualitative}

\begin{figure*}[t]
    \centering
    \includegraphics[width=\textwidth]{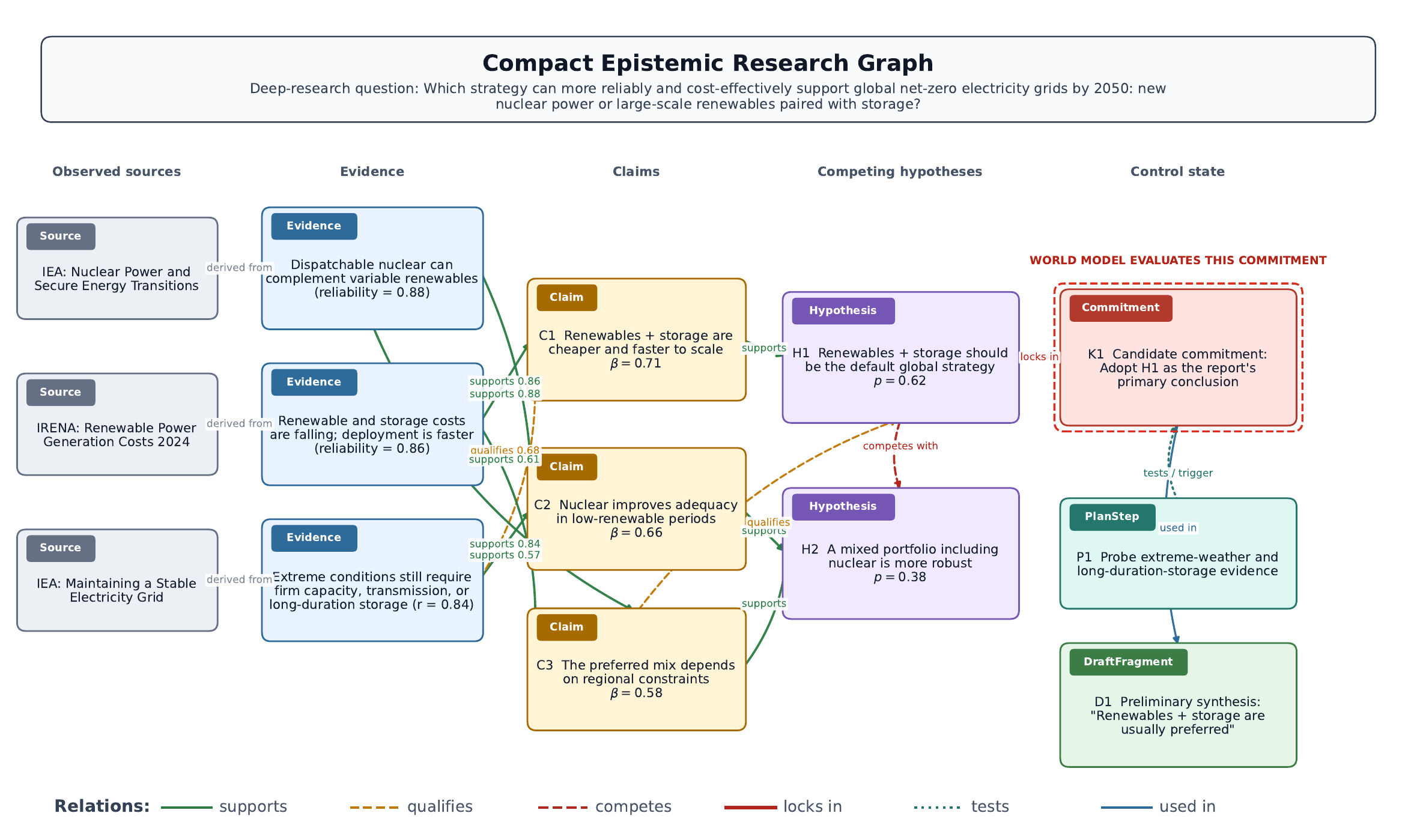}
    \caption{A compact epistemic research graph for a qualitative example. The candidate commitment \(K_1\) would lock the report into \(H_1\), despite remaining support for the competing hypothesis \(H_2\).}
    \label{fig:qualitative_graph}
\end{figure*}

Figure~\ref{fig:qualitative_graph} presents a compact epistemic state for the question of whether new nuclear power or large-scale renewables with storage can more reliably and cost-effectively support global net-zero electricity grids by 2050. The graph preserves the provenance of each finding while exposing the unresolved tension between two competing hypotheses. Evidence on declining renewable and storage costs supports the claim that renewables are cheaper and faster to scale, which in turn supports \(H_1\): renewables with storage should be the default global strategy. However, separate evidence indicates that dispatchable nuclear generation may improve adequacy during low-renewable periods, supporting \(H_2\): a mixed portfolio including nuclear may be more robust. The regional-dependence claim further qualifies any universal conclusion.

\paragraph{World-model prediction.}
The candidate action \(K_1\) commits the report to \(H_1\) as its primary conclusion. Although \(H_1\) currently has higher plausibility than \(H_2\), the difference is not sufficiently decisive: \(H_1\) has probability of \(0.62\), while the alternative retains probability \(0.38\) and is supported by multiple high-reliability evidence nodes. The world model therefore predicts that committing to \(K_1\) would cause substantial hypothesis narrowing by prematurely suppressing \(H_2\). It also identifies non-negligible information loss because the conclusion ``renewables with storage are preferred'' compresses away important qualifications concerning extreme operating conditions and regional constraints. The predicted recovery cost is also elevated because \(K_1\) is directly used in the preliminary draft fragment \(D_1\); later invalidation would therefore require revising both the commitment and the text derived from it.

\paragraph{Recommended control decision.}
Given this prediction, \sysname{} recommends deferring \(K_1\) rather than accepting it immediately. The competing hypotheses remain active, and the agent instead executes plan step \(P_1\), which searches specifically for evidence about extreme-weather reliability and long-duration storage. This plan also defines a concrete contradiction trigger for reconsidering the commitment. If the additional evidence independently supports the adequacy of renewable-storage systems, the agent can later commit to \(H_1\) with lower reversibility risk. If it instead strengthens \(H_2\), the agent can reject \(K_1\) without having propagated it further into the report.

If \(K_1\) had already been accepted, the same graph enables dependency-aware rollback. The repair procedure would retract \(K_1\), mark \(D_1\) as stale because it was derived from that commitment, and preserve the independently supported sources, evidence, claims, and competing hypothesis. Thus, the graph supports both preventive control before commitment and localized repair after contradictory evidence is observed.

\section{Prompts}
\label{app:prompts}

We use three prompts to initialize and update the research-state representation. Prompt~\ref{prompt:target_proposition_elicitation} converts the input research question into a single concise and testable target proposition whose support can be evaluated against subsequently collected evidence. Prompt~\ref{prompt:competing_alternative_elicitation} then elicits mutually distinguishable answer alternatives, forming the initial hypothesis space within which subsequent evidence and commitments are assessed. Finally, Prompt~\ref{prompt:world_model_commitment_prediction} acts as the research-state world model: given a candidate \textsc{commit} action, it predicts the resulting graph changes and estimates their reversibility through hypothesis narrowing, information loss, recovery cost, contradiction-trigger thresholds, and future trigger coverage.

\begin{promptbox*}{Initial Target-Proposition Elicitation}
\label{prompt:target_proposition_elicitation}

Given the research question below, produce exactly one concise and testable target proposition that can be supported or contradicted by evidence.

\textbf{Requirements}

\begin{itemize}
    \item Return exactly one proposition.
    \item Express the proposition as a single sentence.
    \item Ensure that the proposition is concise and unambiguous.
    \item Ensure that the proposition is empirically or logically testable.
    \item Formulate it so that subsequent evidence can either support or contradict it.
    \item Do not include explanations, alternatives, qualifications, or additional text.
\end{itemize}

\textbf{Research question}

\{\{question\}\}

Return only the target proposition.

\end{promptbox*}

\begin{promptbox*}{Competing-Alternative Elicitation}
\label{prompt:competing_alternative_elicitation}

Given the research question below, identify competing answer alternatives that represent distinguishable possible resolutions of the question.

\textbf{Requirements}

\begin{itemize}
    \item Return a JSON array containing at least two alternatives.
    \item Keep each alternative concise.
    \item Ensure that the alternatives are mutually distinguishable.
    \item Avoid returning duplicate or semantically equivalent alternatives.
    \item Do not include explanations, labels, comments, or Markdown fences.
    \item Return valid JSON only.
\end{itemize}

\textbf{Research question}

\{\{question\}\}

Return only the JSON array of competing alternatives.

\end{promptbox*}

\begin{promptbox*}{World-Model Commitment Prediction}
\label{prompt:world_model_commitment_prediction}

You are a world model of a deep-research agent's epistemic state. Given the current research-state graph and a candidate \textsc{commit} action, predict the resulting graph changes and their reversibility.

Do not create, modify, or predict \texttt{Source} or \texttt{Evidence} nodes. You may only reference existing ones by alias. Predict changes only to \texttt{Claim}, \texttt{Hypothesis}, \texttt{Assumption}, \texttt{Commitment}, \texttt{DraftFragment}, and \texttt{PlanStep} nodes. Return one JSON object with:

\begin{itemize}
    \item \texttt{update\_nodes}: created or updated predictable nodes, using
    \texttt{\{"alias":"...","type":"...","op":"update|create","fields":\{...\}\}}.
    
    \item \texttt{update\_edges}: added or removed edges among predictable nodes, or between existing \texttt{Source}/\texttt{Evidence} nodes and predictable nodes, using
    \texttt{\{"op":"add|remove","src":"...","dst":"...",}\\
    \texttt{"relation":"...","weight":0.0\}}.
    
    \item \texttt{reversibility}, containing:
    \begin{itemize}
        \item \texttt{kappa} in $[0,1]$: narrowing of competing hypotheses;
        \item \texttt{lambda} in $[0,1]$: information loss;
        \item \texttt{gamma} in $[0,1]$: normalized recovery cost; and
        \item \texttt{theta}: 
        \texttt{\{"claim\_alias":"...","rho\_star":0.0,"w\_star":0.0\}}, or \texttt{null}.
    \end{itemize}
    
    \item optionally, \texttt{trigger\_coverage} in $[0,1]$: the likelihood that future probes detect contradictions to the committed claim.
\end{itemize}

Use this exact top-level structure:

\begin{quote}
\small\ttfamily
\{ \par
\hspace*{1em}"update\_nodes": [\ldots], \par
\hspace*{1em}"update\_edges": [\ldots], \par
\hspace*{1em}"reversibility": \{ \par
\hspace*{2em}"kappa": 0.44, \par
\hspace*{2em}"lambda": 0.12, \par
\hspace*{2em}"gamma": 0.31, \par
\hspace*{2em}"theta": \{ \par
\hspace*{3em}"claim\_alias": "C1", \par
\hspace*{3em}"rho\_star": 0.75, \par
\hspace*{3em}"w\_star": 0.6 \par
\hspace*{2em}\} \par
\hspace*{1em}\}, \par
\hspace*{1em}"trigger\_coverage": 0.5 \par
\}
\end{quote}

All numeric values must be plain JSON numbers in $[0,1]$. Use existing aliases unless creating a new predictable node, in which case use the appropriate node-type prefix.

\textbf{Current research-state graph}

\{\{research\_state\_graph\}\}

\textbf{Candidate \textsc{commit} action}

\{\{candidate\_commit\_action\}\}

Return only the JSON object, without prose or Markdown fences.

\end{promptbox*}

\end{document}